\documentclass[pmlr]{jmlr}

\usepackage{longtable}

\usepackage{booktabs}
\usepackage{mathtools}
\usepackage[T1]{fontenc}
\DeclarePairedDelimiter\norm{\lVert}{\rVert}

\usepackage[load-configurations=version-1]{siunitx} 

\theorembodyfont{\upshape}
\theoremheaderfont{\scshape}
\theorempostheader{:}
\theoremsep{\newline}

\jmlrvolume{}
\firstpageno{1}
\editors{Sophia Sanborn, Christian Shewmake, Simone Azeglio, Nina Miolane}

\jmlryear{2023}
\jmlrworkshop{Symmetry and Geometry in Neural Representations}

\title[AMES: A Differentiable Embedding Space Selection Framework for Latent Graph Inference]{AMES: A Differentiable Embedding Space Selection Framework for Latent Graph Inference}

 \author{\Name{Yuan Lu} \Email{yl879@cantab.ac.uk}\\
 \addr University of Cambridge, Department of Engineering
 \AND
 \Name{Haitz {S\'aez de Oc\'ariz Borde}} \Email{haitz@oxfordrobotics.institute}\\
 \addr University of Oxford, Oxford Robotics Institute
 \AND
 \Name{Pietro Li\`o} \Email{pl219@cam.ac.uk}\\
 \addr University of Cambridge, Department of Computer Science \& Technology
 }

\begin{document}
\pagenumbering{gobble}

\maketitle

\begin{abstract}
In real-world scenarios, although data entities may possess inherent relationships, the specific graph illustrating their connections might not be directly accessible. Latent graph inference addresses this issue by enabling Graph Neural Networks (GNNs) to operate on point cloud data, dynamically learning the necessary graph structure. These graphs are often derived from a latent embedding space, which can be modeled using Euclidean, hyperbolic, spherical, or product spaces. However, currently, there is no principled differentiable method for determining the optimal embedding space. In this work, we introduce the Attentional Multi-Embedding Selection (AMES) framework, a differentiable method for selecting the best embedding space for latent graph inference through backpropagation, considering a downstream task. Our framework consistently achieves comparable or superior results compared to previous methods for latent graph inference across five benchmark datasets. Importantly, our approach eliminates the need for conducting multiple experiments to identify the optimal embedding space. Furthermore, we explore interpretability techniques that track the gradient contributions of different latent graphs, shedding light on how our attention-based, fully differentiable approach learns to choose the appropriate latent space. In line with previous works, our experiments emphasize the advantages of hyperbolic spaces in enhancing performance. More importantly, our interpretability framework provides a general approach for quantitatively comparing embedding spaces across different tasks based on their contributions, a dimension that has been overlooked in previous literature on latent graph inference.
\end{abstract}


\section{Introduction}

Graph Neural Networks (GNNs) efficiently leverage the geometric prior provided by the edge connections of graph-structured data through the utilization of an adjacency matrix. However, in many real-world applications, the graph connectivity structure may be incomplete, noisy, or even completely unknown. Utilizing latent graph inference enables the application of GNNs to point cloud data without requiring an input adjacency matrix, and it facilitates on-the-fly inference of the optimal graph structure for computation. Numerous studies have employed the Differentiable Graph Module (DGM)~\citep{dDGM} framework to infer latent graphs. This technique hinges on an embedding space, which is employed to deduce distances between nodes within the point cloud. Subsequently, connections are established based on these distances, ultimately shaping a graph. However, the task of selecting the most suitable embedding space is far from straightforward. Various approaches have explored the use of Euclidean space, hyperbolic spaces, spherical spaces, and stereographic projections of these spaces~\citep{borde2023projections}, as well as product spaces~\citep{Manifold-dDGM}. Nevertheless, a principled approach for systematically selecting an optimal embedding space is lacking, particularly in a differentiable fashion. The latent geometry search procedure often relies on random search, which is computationally inefficient, particularly for product manifolds due to the multitude of possible manifold constructions.

In this study, we present an attention-based mechanism for latent graph inference: the Attentional Multi Embedding Selection (AMES) framework. Our approach effectively harnesses multiple latent candidate graphs derived from various latent spaces simultaneously. By employing an attention mechanism, it selects the most optimal latent representations while remaining differentiable. Consequently, this results in an end-to-end framework for the automated selection of latent embedding spaces in the context of latent graph inference. Importantly, we can achieve the same level of performance as existing latent graph inference methods with a single model, eliminating the need for experimental searches to find the optimal latent geometry. Furthermore, we put forward methods for interpretability that enable us to monitor the learning process of the selection mechanism and provide insights into its internal behavior.



\section{Background} \label{sec:2}

\textbf{Geometric Deep Learning and Neural Latent Geometry Search.} Geometric Deep Learning~(GDL) recognizes that certain real-world problems involve data that cannot be adequately represented using simple grids or Euclidean vectors, and aims to generalize traditional machine learning models to a more diverse set of data types, such as point clouds, graphs, and manifolds, to name a few~\citep{GDL_2}. Within the realm of GDL many concepts from differential geometry and topology have been imported into the machine learning community to enhance the performance of learning algorithms~\citep{Bortoli2022RiemannianSG,Hensel2021ASO, Chamberlain2021,Huang2022RiemannianDM}. This includes using manifolds such as the Poincaré ball model~\citep{Mathieu2019}, the hyperboloid~\citep{HGCN}, or the hypersphere~\citep{mettes2019hyperspherical} to encode latent representations of data instead of the Euclidean hyperplane. These manifolds provide a means to represent diverse geometries in a reasonably straightforward and computationally manageable manner. This is possible due to their well-defined mathematical expressions for notions like exponential maps and geodesic distance functions. Despite the promising results obtained from these approaches, a principled method for selecting the appropriate latent embedding manifold construction is still lacking. This has led to the formulation of Neural Latent Geometry Search (NLGS)~\citep{borde2023neural}, which can be formalized as follows: given a search space $\mathfrak{G}$ denoting the set of all possible latent geometries, and the objective function $L_{T,A}(g)$ which evaluates the performance of a given geometry $g$ on a downstream task $T$ for a machine learning model architecture $A$, the objective is to find an optimal latent geometry $g^{*}$: $g^{*}=\textrm{argmin}_{g\in\mathfrak{G}} L_{T,A}(g)$, where in the domain of latent graph inference, $\mathfrak{G}$ represents the set encompassing conceivable geometric similarity measures employed for generating latent graphs.


\textbf{Latent Graph Inference}. Utilizing latent graph inference allows for the integration of GNNs with point cloud data, potentially enhancing performance by harnessing the inductive bias of input graphs or acquiring an entirely new graph from scratch. Some previous contributions have focused on improving existing graph topology by pre-processing the existing adjacency matrices~\citep{bottleneck,GDC,EGP}. The emphasis of this study lies in techniques that dynamically learn and adapt the graph topology without necessitating a rewiring of the initial graph structure. Numerous methods have been put forth to achieve this dynamic learning of the underlying graph topology, in conjunction with the optimization of GNN model parameters: from the original Dynamic Graph Convolutional Neural Network (DGCNN)~\citep{DGCNN} which generates graphs dynamically based on the latent feature activations of the network, to the discrete Differentiable Graph Module (dDGM)~\citep{DGM,dDGM} which performs latent graph inference and message passing in parallel rather than sharing the same embedding space. A comprehensive overview of comparable methods is available in~\cite{Zhu2021DeepGS}, with classical approaches encompassing LDS-GNN~\citep{Franceschi2019LearningDS}, IDGL~\citep{Chen2020IterativeDG}, and Pro-GNN~\citep{Jin2020GraphSL}.

The Differentiable Graph Module (DGM) is a general end-to-end approach for learning latent graphs based on similarity measures between latent node features. Similarity is computed using a distance function that depends on the geometry of the embedding space. Then, based on the distance, the probability of there existing an edge between two nodes is computed, assuming a higher likelihood the closer latent representations of nodes are. Lastly, edge sampling is performed using the Gumbel Top-k trick.  Note that in this work, we focus on the discrete form of the DGM module, the discrete Differentiable Graph Module (dDGM), as dDGM produces sparse graphs which are computationally more efficient and it was also originally recommended over other continuous versions of DGM by~\cite{dDGM}. \cite{Manifold-dDGM} generalized the dDGM module to incorporate non-Euclidean constant curvature hyperbolic and spherical spaces as potential latent graph embedding spaces, as well as Cartesian products of these. This led to a combinatorial space of candidate manifolds to model the dDGM embedding. In this work, we will introduce a differentiable framework to select between potential embeddings.



\vspace{-10pt}
\section{Method: Attentional Multi Embedding Selection for Latent Graph Inference}

\begin{figure}[!htb]    
\centering
\includegraphics[width = 0.67\textwidth]{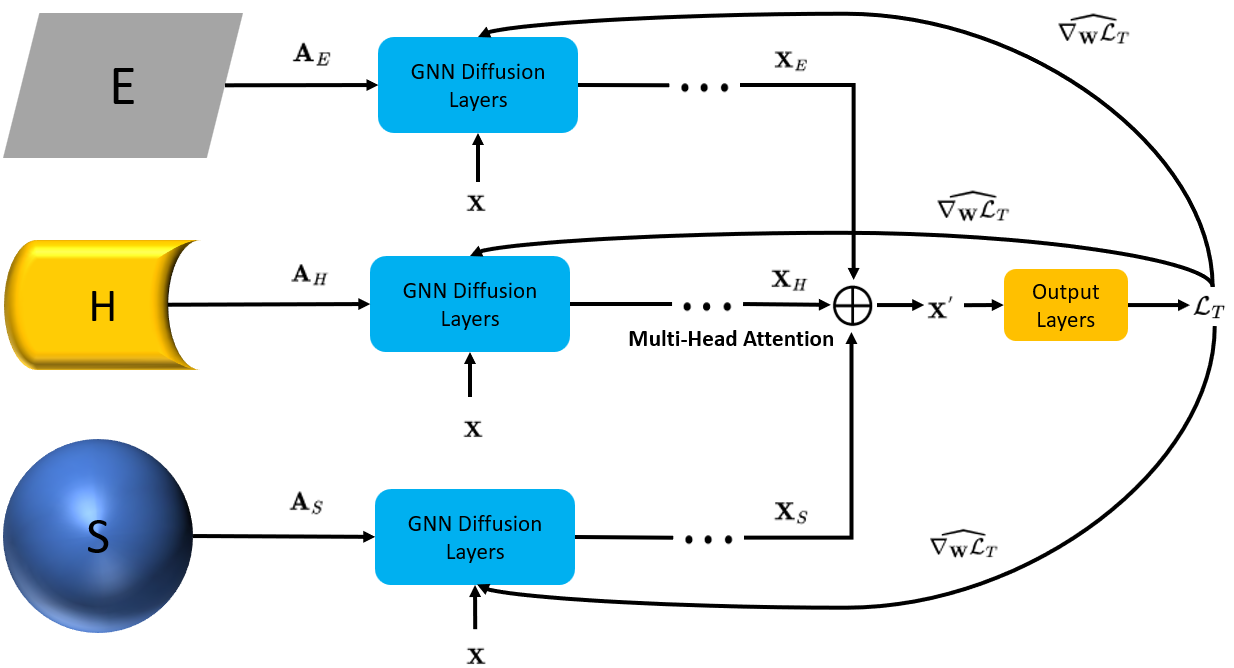}

\caption{\small AMES model depiction. AMES employs multiple dDGM modules running in parallel, each functioning within its distinct embedding space. Consequently, every dDGM generates a unique unweighted adjacency matrix representing its inferred latent graph. Adjacency matrices $\mathbf{A}_{E}, \mathbf{A}_{H}, \mathbf{A}_{S}$ correspond to the Euclidean hyperplane, hyperboloid, and hypersphere models, respectively. These adjacency matrices are then passed to their respective diffusion layers denoted as $g_{\Phi_{\mathcal{M}_{i}}}$. In parallel, new node features are computed based on the different latent graphs. Finally, attention is utilized to combine the features $\mathbf{X}_{\mathcal{M}_{i}}\in[\mathbf{X}_{E},\mathbf{X}_{H},\mathbf{X}_{S}]$ into a unified representation, $\mathbf{X}^{\prime}$. This is subsequently fed into the downstream layers. To update the diffusion layers $g_{\Phi_{\mathcal{M}_{i}}}$ for each dDGM, a shared gradient denoted as \text{\scriptsize$\widehat{\nabla_{\mathbf{W}}\mathcal{L}_{T}}$} is employed.} 
\label{fig:model_arch}               
\vspace{-0.8cm}
\end{figure}

\textbf{Framework Overview.} As mentioned earlier, the primary limitation of existing latent graph inference models is the requirement to predefine the embedding space used to generate the latent graph. This constraint is restrictive because we often lack prior knowledge of which manifold is most suitable for the specific problem we are tackling. Consequently, we aim to introduce a novel framework for differentiable embedding space selection, which we coin Attentional Multi Embedding Selection (AMES). This framework will have the capacity to autonomously learn and select an appropriate embedding space during the training process, providing greater flexibility and adaptability. To do so, we start from a set of candidate latent geometries, as illustrated in Figure~\ref{fig:model_arch}. These candidate latent geometries give rise to distinct latent graphs. These are then input into GNN diffusion layers, which facilitate the propagation of the original node features based on the neighborhoods defined by the suggested latent graphs. Subsequently, we employ self-attention~\citep{NMT,Transformer} to aggregate the new node features, which are then utilized for downstream tasks. Finally, we apply backpropagation based on downstream performance criteria to dynamically determine the optimal embedding space. More precisely, our model learns to assign varying degrees of importance to different candidate latent graphs. Next, we will provide a more detailed explanation of how the feature aggregation and backpropagation processes are executed.

\vspace{5pt}
\textbf{Combining Feature Representations from Candidate Latent Graphs using Attention.} The optimal embedding space for latent graph inference is unknown and varies from task to task. Different embedding spaces come with distinct metrics, resulting in varying latent graphs. Instead of choosing an embedding space in advance, we aim to empower our model to select the best embedding space in a differentiable manner. To achieve this, we will not work with a single dDGM module for latent graph inference but rather employ multiple of them simultaneously. Each dDGM will utilize a different embedding space, suggesting a distinct candidate latent graph. Consequently, this approach will generate different downstream features as information from the upstream layer diffuses through each unique latent graph. Let us denote by $\mathbf{X}_{\mathcal{M}_{i}} \in \mathbb{R}^{N \times d}$ the node feature representations obtained based on different embedding space manifolds $\mathcal{M}_{i}$, where $N$ is the number of nodes in the graph and $d$ the dimensionality of the node features. Next, we define $\mathcal{W}$ as the collection of potential embedding space manifolds for each dDGM with cardinality $M$. Additionally, consider $\mathbf{x}_{k\mathcal{M}_{i}}$ to represent the $k^{th}$ row (or node) of the matrix $\mathbf{X}_{\mathcal{M}_{i}}$. We calculate attention in the following manner:
\begin{equation}
        \alpha_{\mathcal{M}_{i}\mathcal{M}_{j}}^{k}= \frac{exp\left( \frac{\big(\mathbf{W}_{Q}\mathbf{x}_{k\mathcal{M}_{i}}\big)^{T}\mathbf{W}_{K}\mathbf{x}_{k\mathcal{M}_{j}}}{\sqrt{d}}\right)}{\sum_{\mathcal{M}_{j'} \in \mathcal{W}}  exp\left( \frac{\big(\mathbf{W}_{Q}\mathbf{x}_{k\mathcal{M}_{i}}\big)^{T}\mathbf{W}_{K}\mathbf{x}_{k\mathcal{M}_{j'}}}{\sqrt{d}}\right)}.
        \label{weight_scaled_dot_product_rep_attn}
\end{equation}
In this equation, $\mathbf{W}_{Q}$ and $\mathbf{W}_{K}$ represent the query and key weight matrices. Essentially, this equation enables the comparison of node feature representations for a specific node $k$ in the graph across various embedding spaces. The weight for the representation of the $k^{th}$ node within latent space $\mathcal{M}_{i}$, denoted as $\alpha_{\mathcal{M}_{i}}^{k}$, is determined by averaging over all latent spaces:

\begin{equation}
        \alpha_{\mathcal{M}_{i}}^{k} = \frac{\sum_{\mathcal{M}_{j} \in \mathcal{W}}\alpha_{\mathcal{M}_{j}\mathcal{M}_{i}}^{k}}{M}
        \label{Attention_weights_nodewise}
\end{equation}
The coefficient $\alpha_{\mathcal{M}_{i}}^{k}$ can be interpreted as the mean importance of manifold $\mathcal{M}_{i}$ relative to other manifolds for the $k^{th}$ node representation. Finally, the overall combined node representation for the $k^{th}$ node, $\mathbf{x}^{\prime}_{k}$ can be calculated via the weighted sum:
\begin{equation}
        \mathbf{x}^{\prime}_{k} = \sum_{\mathcal{M}_{i} \in \mathcal{W}}\alpha_{\mathcal{M}_{i}}^{k}\mathbf{x}_{k\mathcal{M}_{i}}.
        \label{attn_rep_combination}
\end{equation}
Once this is calculated, the resultant combined node feature matrix $\mathbf{X}^{\prime}$ can be fed into the subsequent downstream GNN layers.
 
\textbf{Attention-based Gradient Descent for
Automatic Latent Space Selection.} Although the attentional combination of feature representations discussed earlier provides valuable insights into selecting latent spaces, relying solely on these weights as straightforward indicators can be problematic if the GNN diffusion layers are not initialized carefully. Several factors, including variations in the initialization and training states of the GNN diffusion layers $g_{\Phi_{\mathcal{M}_{i}}}$ for each latent space $\mathcal{M}_{i}$, can serve as confounding variables that impact the allocation of attention weights to these spaces. To ensure an equitable assessment of different latent spaces throughout the training process, it is crucial to maintain consistent parameters for all the GNN diffusion layers that operate downstream of the parallel dDGM modules. In mathematical terms, our objective is to retain $\Phi_{\mathcal{M}_{1}}=\Phi_{\mathcal{M}_{2}}=\dots=\Phi_{\mathcal{M}_{i}}=\dots$ for all latent spaces $\mathcal{M}_{i} \in \mathcal{W}$ in each training step. This effectively means we assess multiple candidate latent graphs simultaneously during the training of a single GNN, avoiding the need for an ensemble approach. As a result, we can identify the optimal candidate geometry for our model.

The initial step involves ensuring identical parameter initialization for each diffusion layer downstream of every dDGM module. It is important to note that multiple diffusion layers need to be initialized to maintain the integrity of the computational graph. However, in essence, we are evaluating the performance of various candidate latent graphs in parallel on a single model. During each training step, we substitute the initial gradients of the GNN parameters for each latent space, which are obtained through backpropagation, with an attention-based weighted gradient formed by combining individual gradients. We can define $\mathbf{W}_{\mathcal{M}_{i}}$ as the set of model parameters responsible for processing the latent graph generated from the embedding space $\mathcal{M}_{i}$. Let \text{\footnotesize$\nabla_{\mathbf{W}_{\mathcal{M}_{i}}}\mathcal{L}_{T}$} represent the gradient of the loss concerning these parameters for the downstream task. The attention weights for each $\mathcal{M}_{i}$, utilized to combine \text{\footnotesize$\nabla_{\mathbf{W}_{\mathcal{M}_{i}}}\mathcal{L}_{T}$}, are calculated by averaging the node attention weights $\alpha^{k}_{\mathcal{M}_{i}}$, which are computed during the forward pass in line with Equation \ref{Attention_weights_nodewise}:
$\alpha_{\mathcal{M}_{i}} = \frac{\sum_{k=1}^{N}\alpha^{k}_{\mathcal{M}_{i}}}{N},$ where $N$ is the number of nodes. Finally, the attention-weighted gradient shared across all latent spaces is computed as follows:
\begin{equation}
       \widehat{\nabla_{\mathbf{W}}\mathcal{L}_{T}}= \sum_{\mathcal{M}_{i} \in \mathcal{W}} \alpha_{\mathcal{M}_{i}}\Big(\nabla_{\mathbf{W}_{\mathcal{M}_{i}}}\mathcal{L}_{T}\Big).
       \label{attention_weighted_grad}
\end{equation}
This gradient can be employed to update all the parallel diffusion layers using any Stochastic Gradient Descent (SGD)-based optimization method, while ensuring uniform downstream GNN parameters across all latent spaces throughout the training process.
The attention-based feature combination and gradient descent scheme together form our AMES latent space selection framework for embedding space selection.

\textbf{Further Details on dDGM training.} When training GNN models that incorporate dDGM latent graph inference modules, there are two distinct losses to consider. The first one, denoted as $\mathcal{L}_{T}$, assesses the model's performance on downstream tasks, employing cross-entropy loss for classification and mean squared error (MSE) for regression, among other possibilities. Given that the GNN diffuses information using the latent graphs generated by the latent graph inference module, it is important to note that this loss exclusively updates the GNN parameters and does not propagate gradients through the dDGM. Hence, an additional loss term to update the internal learnable parameters of the dDGM is required, which is known as the graph loss, $\mathcal{L}_{GL}$. This loss is based on the negative log likelihood of the probability that an edge exists between two nodes: $\mathcal{L}_{GL}=\sum_{i=1}^{N}\left(\delta(y_{i},\hat{y}_{i})\sum_{l=1}^{L}\sum_{j: (i, j) \in \varepsilon^{(l)}} \log(p_{ij}^{(l)})\right).$ The term $\delta(y_{i},\hat{y}_{i}) = \mathbb{E}(a_{i})-a_{i}$ is characterized as the disparity between two aspects: the average accuracy observed for the $i^{th}$ node and the correctness of the current prediction for that node's label, represented as $a_{i}$. Here, $a_{i}$ equals 1 if $y_{i}=\hat{y}_{i}$ and 0 otherwise, where $y_{i}$ is the label prediction and $\hat{y}_{i}$ the ground truth label. For regression tasks, the R2 score could be used in place of accuracy. In turn, the probability of an edge existing between two given nodes can be expressed as follows, $p_{ij}^{(l)}(\Theta^{(l)})=\exp\left(-T\mathfrak{d}_{\mathcal{M}}\left(\overline{f_{\Theta}^{(l)}(\mathbf{x}_{i}^{(l)})}, \overline{f_{\Theta}^{(l)}(\mathbf{x}_{j}^{(l)})}\right)\right),$ where $T$ is a temperature parameter, $\mathfrak{d}_{\mathcal{M}}$ is the geodesic distance function for a given embedding space manifold $\mathcal{M}$, $\mathbf{x}_{i}^{(l)}$ and $\mathbf{x}_{j}^{(l)}$ are the node feature representations of nodes $i$ and $j$ in layer $l$, and $\overline{f_{\Theta}^{(l)}(\mathbf{x}_{i}^{(l)})}$ signifies that the node features are passed through a parametrized function $f_{\Theta}$ and later projected $\overline{(\cdot)}$ using the exponential map onto the corresponding manifold $\mathcal{M}$. For our attention-based model, the overall graph loss can be computed
by simply adding up individual graph losses from each dDGM module:
\begin{equation}
\mathcal{L}_{GL}^{\prime}=\sum_{\mathcal{M}_{i} \in \mathcal{W}}\mathcal{L}_{GL}^{(\mathcal{M}_{i})}.
\label{attn_graph_loss}
\end{equation}
It is worth noting that while latent graph inference could theoretically be applied to all layers of the GNN, practical experiments, as demonstrated in~\cite{dDGM} and~\cite{Manifold-dDGM}, reveal that stacking multiple dDGM modules does not yield performance improvements, especially when considering the computational cost involved. Consequently, in our work, we will exclusively employ a single dDGM module for all experiments. During the training of AMES, we optimize both $\mathcal{L}_{T}$ and $\mathcal{L}_{GL}^{\prime}$ using the Adam optimizer~\citep{Adam}. To ensure a fair comparison with prior work, we adopt the same number of training epochs and GNN parameterizations as in~\cite{Manifold-dDGM}. Specific details regarding the learning rate and weight decay can be found in Appendix~\ref{appendix:training_stats}. For additional model architecture details refer to Appendix~\ref{appendix:homo_arch},~\ref{appendix:hetero_arch} and~\ref{appendix:TadPole_arch}. 



\section{Training Interpretability Framework}
In addition to numerical outcomes, we employ a gradient-based attribution approach, drawing inspiration from~\cite{CNN_visual}, to assess the significance of each latent space's influence on the downstream tasks. This serves as an interpretability analysis. The saliency maps suggested in~\cite{CNN_visual} are generated by calculating the gradient of the score function of the label class of an input image with respect to the pixels of the input image itself. In an intuitive sense, the gradient of a predicted label in relation to an input image measures how the model's prediction changes when individual image pixels undergo slight modifications. Therefore, a larger gradient signifies a greater influence on the predictions. In our context, the score function is the cross-entropy loss for all the nodes in the graph. Hence, we take the gradient of the loss $\mathcal{L}_{T}$ with respect to $\mathbf{X}_{\mathcal{M}_{i}} \in \mathbb{R}^{N \times d}$, the feature activations produced from individual latent graphs originating in distinct latent spaces, denoted as $\mathcal{M}_{i}$. This returns a gradient matrix $\nabla_{\mathbf{X}_{\mathcal{M}_{i}}}\mathcal{L}_{T} \in \mathbb{R}^{N \times d}$ since the loss is a scalar. Given that $\nabla_{\mathbf{X}_{\mathcal{M}_{i}}}\mathcal{L}_{T}$ is high-dimensional and hard to visualize, we calculate its Frobenius norm:
\begin{equation}
        \norm*{\nabla_{\mathbf{X}_{\mathcal{M}_{i}}}\mathcal{L}_{T}}_{\mathcal{F}} = \sqrt{\sum_{i=1}^{N}\sum_{j=1}^{d} \Big(\nabla_{\mathbf{X}_{\mathcal{M}_{i}}}\mathcal{L}_{T}\Big)_{ij}^{2}}
        \label{Grad_Intepretability}
\end{equation}
which gives a scalar indicator of each latent space's attribution to the node classification prediction and it is easier to track and visualize during training. During training initialization, all representations stemming from diverse latent embedding spaces are set to exert equal influence on the downstream representations, resulting in identical Frobenius norms. As training progresses, the attention coefficients begin to learn which latent graphs are more suitable for computation. By monitoring the Frobenius norm of the loss gradient, we can gauge the pace at which this adaptation occurs and identify which embedding spaces assume a dominant role in the learning process. We do notice the existence of other GNN interpretability approaches \citep{gnnexplainer,pgexplainer,causalscreening}, however, these approaches mainly focusing on
determining subgraphs that are important for individual predictions and hence are not suitable for our task.

\section{Results}

\textbf{Experimental Setup and Numerical Evaluation.} In this work, we focus on transductive node classification. We test AMES on five homophilic, heterophilic and real-world datasets. For each dataset, we perform 10-fold cross-validation where $90\%$ of the nodes are used for training and $10\%$ for testing in each split in line with~\cite{dDGM}. In this case, the task-specific loss is the cross-entropy loss. In our approach, we exclusively employ model spaces as candidate embedding spaces, which suffices to match the performance of prior methods that utilize product manifolds. It is worth noting that for homophilic graph benchmarks, we leverage the original dataset graph as an inductive bias. However, when dealing with heterophilic datasets, we have observed that starting from a point cloud yields more significant benefits. This preference arises from the fact that we perform diffusion with GCN layers. In the instance of TadPole, a real-world baseline dataset, we encounter a unique scenario where no initial input graph is available. Consequently, our latent graph inference system operates directly on the point cloud in this case. Table \ref{table:asel_attn_no_inductive_bias_model_spaces_main} displays the numerical results. AMES consistently demonstrates improved or comparable test set accuracy across various datasets compared to earlier models, even when those models employ more intricate embedding structures, such as product spaces. Notably, the H+S combination consistently surpasses all other models on nearly all five datasets. The key takeaway is that using our proposed model eliminates the need for random exploration among latent spaces.

\begin{table}[!ht]
\caption{\small Accuracy $(\%)$ $\pm$ stdev. \textcolor{red}{First}, \textcolor{blue}{Second} and \textcolor{violet}{Third} are highlighted. k is the number of edges sampled per node when using Gumbel Top-k trick. The homophilic graphs (Cora and CiteSeer) use the dataset graph as inductive bias, the rest do not. E, H, S denote Euclidean, hyperbolic and spherical spaces respectively. Note that, for example, $EH$ denotes that the embedding space is a product, whereas $E+H$ denotes that AMES uses both model spaces simultaneously, not a product.} 
\centering

\resizebox{\textwidth}{!}{
\begin{tabular}{ccccccccc}
    \toprule
     & \multicolumn{2}{c}{\textbf{HOMOPHILIC DATASETS}} &  &\multicolumn{2}{c}{\textbf{HETEROPHILIC DATASETS}} & &\textbf{REAL-WORLD DATASETS}\\
      \midrule
      &  \textbf{Cora} & \textbf{CiteSeer}  &  &\textbf{Chameleon} & \textbf{Squirrel} &  &\textbf{TadPole}\\ 
\multicolumn{1}{l}{Connections k} &7 & 5 &  &2 & 3 & &3\\
    \midrule
    \multicolumn{1}{l}{\textbf{Baseline Models}} \\
    \midrule
    \multicolumn{1}{l}{dDGM-E} & $81.89 \pm 4.05$ & \textcolor{violet}{$73.52 \pm 2.23 $} &  &  $44.93 \pm 2.42$ &  $33.17 \pm 2.11 $& &$83.21 \pm 14.04$\\
    \multicolumn{1}{l}{dDGM-H}  & $82.41 \pm 3.25 $ & $73.22 \pm 1.50 $& &$48.33 \pm 3.45$ & $33.29 \pm 1.90 $& &\textcolor{violet}{$86.96 \pm  9.48$}\\
    \multicolumn{1}{l}{dDGM-S} & $78.59 \pm 4.82 $ & $70.27 \pm 5.05 $ & & $43.88 \pm 3.79$ & $33.04 \pm 1.71$& &$81.07 \pm 12.38$\\ 
    \multicolumn{1}{l}{dDGM-ES} & $74.85 \pm 16.53$ & $71.08 \pm 7.00 $ &   &$47.71 \pm 2.11$ &  \textcolor{violet}{$33.56 \pm 2.39 $} & &$70.71 \pm 17.22$\\
    \multicolumn{1}{l}{dDGM-EH} &\textcolor{red}{$84.74 \pm 4.89 $} & \textcolor{blue}{$73.55 \pm 2.35 $} & &$46.78 \pm 3.16$ & \textcolor{blue}{$33.65 \pm 1.60 $} & &\textcolor{violet}{$86.96 \pm 9.45$}\\
    \multicolumn{1}{l}{dDGM-HS} & \textcolor{violet}{$84.48 \pm 4.55 $} & $67.68 \pm 19.12 $ &  &\textcolor{violet}{$48.41 \pm 2.86$} & $32.87 \pm 1.70$ & &$77.68 \pm 15.72$\\
    \multicolumn{1}{l}{dDGM-EHH} & $79.63 \pm 18.23 $ & $68.70 \pm 16.36 $ & & $48.28 \pm 3.33$ & $33.23 \pm 2.66$ & &$86.43 \pm 10.01$\\
    \multicolumn{1}{l}{dDGM-EHS} & \textcolor{blue}{$84.67 \pm 6.36 $} & $68.83 \pm 16.14 $ &  &\textcolor{blue}{$48.63 \pm 2.30$} & $33.31 \pm 1.95$ & &$83.93 \pm 14.15$\\
    \multicolumn{1}{l}{MLP} & $59.52 \pm 3.67$ & $58.37 \pm 3.29$ &  &$42.86 \pm 2.72 $ & $31.12 \pm 1.99$& &$83.57 \pm 6.12$\\ 
    \multicolumn{1}{l}{GCN} & $81.41 \pm 10.45$ & $70.84 \pm 2.80$  & &  $34.05 \pm 4.03 $ & $24.88 \pm 2.78$ & &N/A\\
    \midrule
    \multicolumn{1}{l}{\textbf{New Models (AMES)}}\\
    \midrule
    \multicolumn{1}{l}{AMES-H+S} &  $83.33 \pm 5.06$ & \textcolor{red}{$73.64 \pm 2.45$} &    &\textcolor{red}{$49.30 \pm 2.33$} & \textcolor{red}{$34.85 \pm 2.02$}& &\textcolor{red}{$90.18 \pm 4.01$}\\
    \multicolumn{1}{l}{AMES-E+H+S} &  $83.78 \pm 3.08$ & $73.43 \pm 2.23$ & & $48.02 \pm 3.82$ & $33.00 \pm 1.44$& &\textcolor{blue}{$87.14 \pm 6.81$}\\
    \bottomrule
\end{tabular}
}
\label{table:asel_attn_no_inductive_bias_model_spaces_main}
\end{table}

\begin{figure}[!ht]
\floatconts
  {fig:e_h_s_no_inductive_bias_grad}
  {\caption{\small Gradient contribution for each model space in AMES during training. We plot the average Frobenius norm over 10 runs against the training epochs for Cora, CiteSeer, Squirrel, Chameleon, and TadPole. We consistently observe that AMES gives the most importance to the hyperbolic embedding space.}}
  {%
    \subfigure[Cora]{\label{fig:cora_e_h_s_no_inductive_bias_grad}%
      \includegraphics[width=0.41\linewidth]{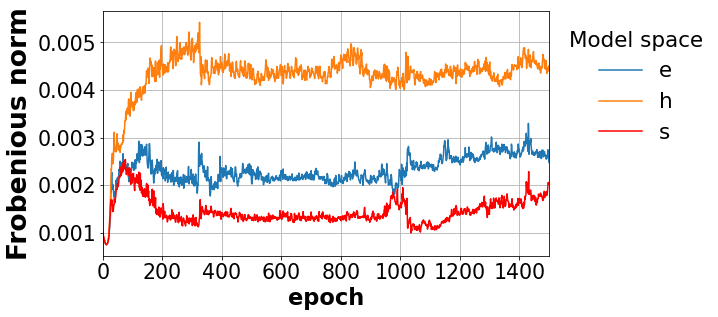}}%
    \qquad
    \subfigure[CiteSeer]{\label{fig:citeseer_e_h_s_no_inductive_bias_grad}%
      \includegraphics[width=0.41\linewidth]{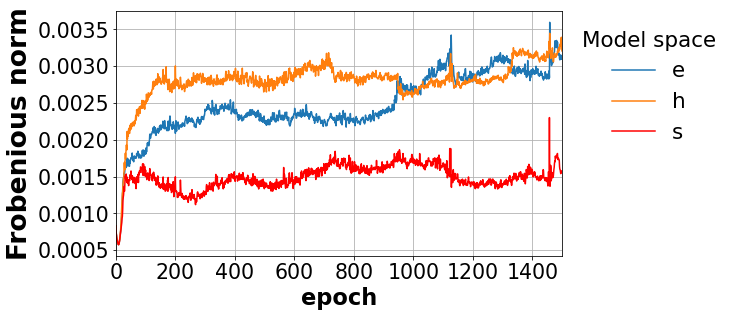}}%
      \qquad
    \subfigure[Squirrel]{\label{fig:squirrel_e_h_s_no_inductive_bias_grad}%
      \includegraphics[width=0.41\linewidth]{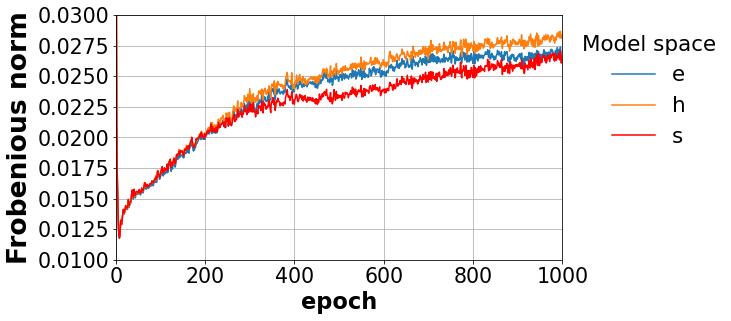}}%
      \qquad
      \subfigure[Chameleon]{\label{fig:chameleon_e_h_s_no_inductive_bias_grad}%
      \includegraphics[width=0.41\linewidth]{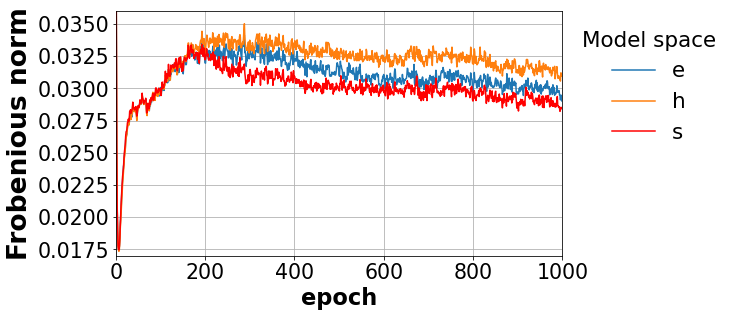}}%
      \qquad
      \subfigure[TadPole]{\label{fig:TadPole_e_h_s_no_inductive_bias_grad}%
      \includegraphics[width=0.47\linewidth]{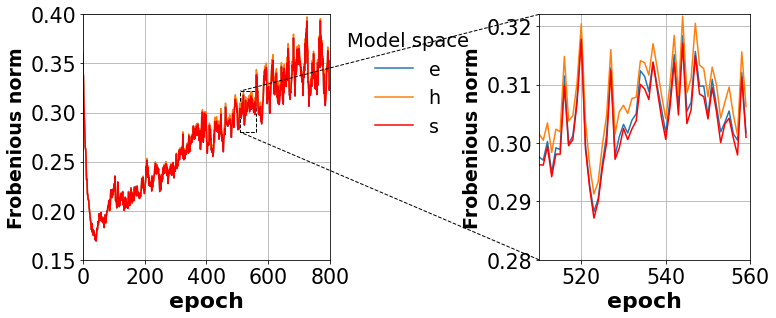}}%
  }
\end{figure}

\textbf{Interpretability Results.} We create visualizations to track the average norm evolution across the 10 training folds. In the interest of conducting the interpretability analysis, despite the superior performance of H+S in principle, Figure~\ref{fig:e_h_s_no_inductive_bias_grad} presents plots illustrating the average Frobenius norm of loss gradients for E+H+S. This allows us to monitor the learning progress across all model spaces, as opposed to focusing solely on two of them. The trends align with the data presented in Table~\ref{table:asel_attn_no_inductive_bias_model_spaces_main} for all five datasets. In the case of Cora, it is noteworthy that $\mathbb{H}$ appears to contribute the most, while $\mathbb{S}$ makes the smallest contribution. For CiteSeer, although the gap between $\mathbb{E}$ and $\mathbb{H}$ is not as pronounced, both curves conform to the baseline results for single model spaces found in Table~\ref{table:asel_attn_no_inductive_bias_model_spaces_main}. For the heterophilic datasets we observe a similar trend, the models using hyperbolic space are superior and our framework learns to select the graph stemming from hyperbolic space as the main contributor, see Figure~\ref{fig:chameleon_e_h_s_no_inductive_bias_grad} and~\ref{fig:squirrel_e_h_s_no_inductive_bias_grad}. In summary, our model demonstrates its ability to autonomously identify the embedding spaces that align most favorably with the baseline results. The attention weights assigned to each latent space closely correspond to their counterparts based on single model spaces in the baselines. As a result, our AMES effectively assesses the relative importance of the involved latent spaces based on their impact on downstream tasks. Furthermore, it becomes evident that the hyperbolic space $\mathbb{H}$ consistently emerges as the preferred choice across the majority of the benchmark graph datasets we have employed. This intriguing finding may merit further investigation beyond the scope of this study and seems to align with the common belief that hyperbolic space is better suited for encoding latent tree structures~\citep{kratsios2023capacity}, such as those present in our graph benchmarks.

\section{Conclusion}

In this work we have introduced AMES, a fully-differentiable mechanism designed for embedding space selection within the context of latent graph inference. In contrast to prior methods, our approach circumvents the need for conducting random searches across the combinatorial spectrum of potential product manifold combinations. Instead, it possesses the capacity to dynamically learn the optimal embedding space through a single model. Moreover, we offer interpretability techniques to monitor this learning process. Notably, our method maintains performance levels on par with other models in the existing literature, all while alleviating the computational burden associated with conducting multiple experiments in pursuit of the ideal embedding space.

\textbf{Future Work.} Our model selection remains grounded in constant curvature model spaces, implying that local curvature control remains a challenge. While the application of attention could potentially yield more flexible representations than what constant curvature model spaces can offer on their own, additional research is required to gain a deeper understanding of how this approach influences the curvature of the embedding space. Note that since the method does not distinctly select a candidate embedding space but rather combines model spaces using attention, the model is effectively learning a new metric.
\bibliography{main}

\appendix 
\section{Training Statistics and Model Architecture}
This section provides the training statistics and model architectures for both the dDGM and the downstream GNN diffusion layer applied to homophilic, heterophilic, and real-world datasets. In general, we closely follow the model architecture and training statistics outlined in~\citet{Manifold-dDGM}, with only minor differences.
\subsection{Training Statistics} \label{appendix:training_stats}

\begin{table}[!ht]
\centering
\caption{The Training statistics of the 5 homophilic, heterophilic and real-world benchmark datasets used in this work for the baseline dDGM, MLP and GCN models.}
\begin{tabular}{cccccc} \hline
  & Training Epochs & Learning Rate & Weight Decay \\ \hline
 \textbf{Cora} & 1,500 & $1\times10^{-2}$& $1\times10^{-4}$\\
 \textbf{CiteSeer} & 1,500 & $1\times10^{-2}$& $1\times10^{-4}$\\
 \textbf{Squirrel} &1,000 &$1\times10^{-2}$ & $1\times10^{-3}$\\
\textbf{Chameleon} & 1,000& $1\times10^{-2}$& $1\times10^{-3}$\\
 \textbf{TadPole} & 800& $1\times10^{-3}$& $2\times10^{-4}$\\
\hline
\end{tabular}
\label{table: Training_statistics_baseline}
\end{table}
\vspace{-8pt}
\begin{table}[!ht]
\centering
\caption{The Training statistics of the 5 homophilic, heterophilic and real-world benchmark datasets used in this work for AMES in this work.}
\begin{tabular}{cccccc} \hline
  & Training Epochs & Learning Rate & Weight Decay \\ \hline
 \textbf{Cora} & 1,500 & $5\times10^{-3}$& $1\times10^{-5}$\\
 \textbf{CiteSeer} & 1,500 & $4\times10^{-3}$& $1\times10^{-5}$\\
 \textbf{Squirrel} &1,000 &$1\times10^{-2}$ & $1\times10^{-5}$\\
\textbf{Chameleon} & 1,000& $1\times10^{-2}$& $1\times10^{-5}$\\
 \textbf{TadPole} & 800& $1\times10^{-3}$& $2\times10^{-4}$\\
\hline
\end{tabular}
\label{table: Training_statistics_new_model}
\end{table}

\newpage
\subsection{Model Architecture for Homophilic Datasets} \label{appendix:homo_arch}

\begin{table}[!ht]
\centering
\caption{Model architecture of dDGM  for homophilic datasets Cora and CiteSeer.}
\vspace{12pt}
\begin{tabular}{lcccc}
    \toprule
     Layer size&    & dDGM \\ 
   \midrule
    \multicolumn{1}{l}{(No. features, 32) } &  &  Linear\\
    N/A &   & ELU\\
    \multicolumn{1}{l}{(32, 16 per model space) } &  & GCN Conv\\
    N/A &   & ELU\\
    \multicolumn{1}{l}{(16 per model space, 4 per model space) } &  & GCN Conv\\
   N/A &  & Sigmoid\\
    \bottomrule
\end{tabular}

\label{table:dDGM_arch_homo}
\end{table}

\begin{table}[!htbp]
\centering
\caption{Model architecture of downstream GNN layers as well as MLP baseline for homophilic datasets Cora and CiteSeer.}
\begin{tabular}{lcccc}
    \toprule
     Layer size&   &MLP & GCN & GCN-dDGM \\ 
   \midrule
   \multicolumn{1}{l}{ } &  & N/A &  N/A & dDGM\\ \hline 
    \multicolumn{1}{l}{(No. features, 32) } &  & Linear &  GCN Conv & GCN Conv\\
    N/A &  & ELU & ELU &ELU\\
    \multicolumn{1}{l}{(32, 16)} &  & Linear& GCN Conv & GCN Conv\\
    N/A &  & ELU & ELU &ELU\\
    \multicolumn{1}{l}{(16, 8) } &  & Linear& GCN Conv & GCN Conv\\
   N/A &  & ELU& ELU &ELU\\
   \multicolumn{1}{l}{(8, 8) } &  & Linear& Linear & Linear\\
   N/A &  & ELU& ELU &ELU\\
   \multicolumn{1}{l}{(8, 8) } &  & Linear& Linear & Linear\\
   N/A &  & ELU& ELU &ELU\\
   \multicolumn{1}{l}{(8, No.classes) } &  & Linear& Linear & Linear\\
    \bottomrule
\end{tabular}
\label{table: GNN_arch_homo}
\end{table}
\newpage
\subsection{Model Architecture for Heterophilic Datasets} \label{appendix:hetero_arch}

\begin{table}[!ht]
\centering
\caption{Model architecture of dDGM for heterophilic datasets Squirrel and Chameleon.}
\vspace{12pt}
\begin{tabular}{lcccc}
    \toprule
     Layer size&   &dDGM \\ 
   \midrule
    \multicolumn{1}{l}{(No. features, 32) } &  & Linear\\
    N/A &  & BatchNorm \\
    N/A &  & ELU \\
    \multicolumn{1}{l}{(32, 4 per model space) } &  & Linear\\
    N/A &  & BatchNorm \\
    N/A &  & ELU \\
    \multicolumn{1}{l}{(4 per model space, 4 per model space) } &  & Linear\\
    N/A &  & BatchNorm \\
   N/A &  & Sigmoid\\
    \bottomrule
\end{tabular}

\label{table:dDGM_arch_hetero}
\end{table}

\begin{table}[!htbp]
\centering
\caption{Model architecture of downstream GNN layers as well as MLP baseline for heterophilic datasets Squirrel and Chameleon.}
\begin{tabular}{lcccc}
    \toprule
     Layer size&   &MLP  & GCN-dDGM \\ 
   \midrule
   \multicolumn{1}{l}{ } &  & N/A  & dDGM\\ \hline 
    \multicolumn{1}{l}{(No. features, 16) } &  & Linear  & GCN Conv\\
    N/A &  & ELU  &ELU\\ 
    \multicolumn{1}{l}{(16, 8) } &  & Linear & GCN Conv\\
   N/A &  & ELU &ELU\\
   \multicolumn{1}{l}{(8, 8) } &  & Linear & Linear\\
   N/A &  & BatchNorm &BatchNorm\\
   N/A &  & ELU &ELU\\
   \multicolumn{1}{l}{(8, No.classes) } &  & Linear & Linear\\
    \bottomrule
\end{tabular}
\label{table: GNN_arch_hetero}
\end{table}

\newpage
\subsection{Model Architecture for TadPole} \label{appendix:TadPole_arch}

\begin{table}[!ht]
\centering
\caption{Model architecture of dDGM for TadPole.}
\vspace{12pt}
\begin{tabular}{lcccc}
    \toprule
     Layer size&   &dDGM \\ 
   \midrule
    \multicolumn{1}{l}{(No. features, 16 per model space) } &  & Linear\\
    N/A &  & BatchNorm \\
    N/A &  & ELU \\
    \multicolumn{1}{l}{(16 per model space, 4 per model space) } &  & Linear\\
    N/A &  & BatchNorm \\
   N/A &  & Sigmoid\\
    \bottomrule
\end{tabular}
\label{table:dDGM_arch_TadPole}
\end{table}

\begin{table}[!htbp]
\centering
\caption{Model architecture of downstream GNN layers as well as MLP baseline for TadPole.}
\begin{tabular}{lcccc}
    \toprule
     Layer size&   &MLP  & GCN-dDGM \\ 
   \midrule
   \multicolumn{1}{l}{ } &  & N/A  & dDGM\\ \hline 
    \multicolumn{1}{l}{(No. features, 32) } &  & Linear  & GCN Conv\\
    N/A &  & ELU  &ELU\\
    \multicolumn{1}{l}{(32, 16) } &  & Linear & GCN Conv\\
   N/A &  & ELU &ELU\\
   \multicolumn{1}{l}{(16, 8) } &  & Linear & GCN Conv\\
   N/A &  & ELU &ELU\\
   \multicolumn{1}{l}{(8, 8) } &  & Linear & Linear\\
   N/A &  & BatchNorm &BatchNorm\\
   N/A &  & ELU &ELU\\
   \multicolumn{1}{l}{(8, 8) } &  & Linear & Linear\\
   N/A &  & BatchNorm &BatchNorm\\
   N/A &  & ELU &ELU\\
   \multicolumn{1}{l}{(8, No.classes) } &  & Linear & Linear\\
    \bottomrule
\end{tabular}
\label{table: GNN_arch_TadPole}
\end{table}

\end{document}